\pdfoutput=1

\documentclass[11pt]{article}

\usepackage[preprint]{acl}
\usepackage{times}
\usepackage{latexsym}
\usepackage[T1]{fontenc}
\usepackage[utf8]{inputenc}
\usepackage{xcolor}
\usepackage{tcolorbox}
\usepackage{microtype}
\usepackage{inconsolata}
\usepackage{graphicx}
\usepackage{subcaption}
\usepackage{booktabs}
\usepackage{multirow}
\usepackage{tablefootnote}
\usepackage{hyperref}
\definecolor{myblue}{RGB}{150, 150, 230}

\usepackage{algorithm}
\usepackage{algorithmic}
\usepackage{booktabs}
\usepackage{multirow}
\usepackage{xcolor}
\usepackage{colortbl}
\usepackage{amssymb}
\usepackage{pifont}

\usepackage{amsmath,amsfonts,bm}

\def\eqref#1{equation~\ref{#1}}

\def\1{\bm{1}}

\def\vp{{\bm{p}}}

\DeclareMathAlphabet{\mathsfit}{\encodingdefault}{\sfdefault}{m}{sl}
\SetMathAlphabet{\mathsfit}{bold}{\encodingdefault}{\sfdefault}{bx}{n}

\def\gL{{\mathcal{L}}}

\DeclareMathOperator*{\Ours}{UMed-LVLM}

\title{
Improving Medical Large Vision-Language Models \\with Abnormal-Aware Feedback
}

\author{
Yucheng Zhou, 
Lingran Song, 
Jianbing Shen\thanks{~Corresponding author. This work was supported by the National Natural Science Foundation of China (No. 624B2002) and the Jiangyin Hi-tech Industrial Development Zone under the Taihu Innovation Scheme (EF2025-00003-SKL-IOTSC).}\\
SKL-IOTSC, CIS, University of Macau \\
{\tt yucheng.zhou@connect.um.edu.mo, jianbingshen@um.edu.mo}
}
    
\begin{document}
\maketitle

\begin{abstract}
Existing Medical Large Vision-Language Models (Med-LVLMs), encapsulating extensive medical knowledge, demonstrate excellent capabilities in understanding medical images. However, there remain challenges in visual localization in medical images, which is crucial for abnormality detection and interpretation. 
To address these issues, we propose a novel UMed-LVLM designed to unveil medical abnormalities. 
Specifically, we collect a Medical Abnormalities Unveiling (MAU) dataset and propose a two-stage training method for UMed-LVLM training. 
To collect MAU dataset, we propose a prompt method utilizing the GPT-4V to generate diagnoses based on identified abnormal areas in medical images. 
Moreover, the two-stage training method includes Abnormal-Aware Instruction Tuning and Abnormal-Aware Rewarding, comprising Relevance Reward, Abnormal Localization Reward and Vision Relevance Reward. 
Experimental results demonstrate that our UMed-LVLM significantly outperforms existing Med-LVLMs in identifying and understanding medical abnormalities, achieving a 58\% improvement over the baseline.
In addition, this work shows that enhancing the abnormality detection capabilities of Med-LVLMs significantly improves their understanding of medical images and generalization capability. 
\end{abstract}

\section{Introduction}
Large Vision-Language Models (LVLMs) demonstrate remarkable capability in various vision-language tasks \cite{openai2023gpt4,liu2023visual,LLaVA-Med}. Medical image analysis poses a significant challenge for LVLMs due to their intricate patterns and structures, thereby demanding an in-depth grasp of nuanced variations to ensure precise diagnoses \cite{wu2023can}. To enhance LVLMs for medical images, some works \cite{DBLP:conf/iclr/QinYL023} encapsulate a substantial medical corpus into these models, i.e., Medical Large Vision-Language Models (Med-LVLMs). These Med-LVLMs exhibit proficiency in understanding medical images and human queries.

\begin{table}[!t]\small
\centering
\setlength{\tabcolsep}{2.5pt}
\resizebox{\linewidth}{!}{
\begin{tabular}{lcccc}
\toprule
\multirow{2}{*}{\textbf{Method}} & \multicolumn{1}{c}{\textbf{Medical}} & \multicolumn{1}{c}{\textbf{Multi-modal}} & \multicolumn{1}{c}{\textbf{Region-}} & \multicolumn{1}{c}{\textbf{Detector $\&$}} \\ 
& \multicolumn{1}{c}{\textbf{Diagnosis}} & \multicolumn{1}{c}{\textbf{Medical Image}} & \multicolumn{1}{c}{\textbf{Awareness}} & \multicolumn{1}{c}{\textbf{Segmenter-Free}} \\ 
\midrule
RegionGPT      & \ding{55} & \ding{55} & \ding{51} & \ding{55} \\ 
XrayGPT        & \ding{51} & \ding{55} & \ding{55} & \ding{51} \\ 
Med-Flamingo   & \ding{51} & \ding{51} & \ding{55} & \ding{51} \\ 
MedVInt        & \ding{51} & \ding{51} & \ding{55} & \ding{51} \\ 
\rowcolor{gray!15}Our UMed-LVLM      & \ding{51} & \ding{51} & \ding{51} & \ding{51} \\ 
\bottomrule
\end{tabular}
}
\caption{\small Comparison of different Med-LVLMs: RegionGPT \cite{DBLP:journals/corr/abs-2403-02330}, XrayGPT \cite{thawkar2023xraygpt}, Med-Flamingo \cite{DBLP:conf/ml4h/MoorHWYDLZRR23}, MedVInt \cite{DBLP:journals/corr/abs-2305-10415} and Ours. ``Medical Diagnosis'' denotes the model's applicability for medical diagnosis; ``Multi-modal Medical Image'' indicates training on multi-modal medical images; ``Region-Awareness'' reflects the model's capability for region recognition; ``Detector \& Segmenter-Free'' specifies independence from external detectors or segmenters.}
\label{tab:comparison}
\end{table}

Despite their successes, existing Med-LVLMs exhibit limitations in visual localization capability within medical images, as shown in Table~\ref{tab:comparison}. Advanced models like GPT-4V \citep{openai2023gpt4}, one of the leading LVLMs, exhibit notable shortcomings in accurately interpreting and visual localization in medical images (i.e., abnormality localization) \cite{wu2023can}. The capability for visual localization is critical for two primary reasons: Firstly, bias in visual localization can lead to unreliable responses in diagnosis, undermining the credibility and interpretability of Med-LVLMs. Secondly, some LVLMs~\cite{DBLP:journals/corr/abs-2310-16045,DBLP:journals/corr/abs-2403-02330,LocalizeObject} improve the visual understanding in LVLM by enhancing its visual localization capability. In these works, visual localization in natural scenes can benefit from general detectors. However, abnormality localization in medical images lacks a large amount of data to train specialized detectors, especially for some rare diseases. This limitation emphasizes the necessity for enhancing the inherent visual localization capabilities of medical LVLMs without specific detectors to improve their understanding of medical images and the reliability of responses.

To address these challenges, we propose \textbf{UMed-LVLM}, designed with \textbf{U}nveiling \textbf{Med}ical abnormalities for \textbf{LVLM}. To encapsulate the abnormality unveiling capability into UMed-LVLM, we first collect a dataset comprising medical images with abnormality regions. Then, we design a prompt method to generate a diagnosis dataset, i.e., Medical Abnormalities Unveiling (MAU), through the GPT-4V model \citep{2023GPT4VisionSC}. 
The MAU dataset encompasses 5,817 medical images, user queries, and diagnosis responses with abnormal areas. 
This dataset is used to train UMed-LVLM via a two-stage training method, i.e., Abnormal-Aware Instruction Tuning and Abnormal-Aware Rewarding, enabling it to understand medical abnormalities. 
UMed-LVLM is continually trained on MedVInt~\cite{DBLP:journals/corr/abs-2305-10415}. 
To prevent knowledge catastrophic forgetting of large language models during continual training on large-scale datasets~\cite{Forgetting}, we train the model on limited-scale datasets, following that employed in previous studies~\cite{LIMA,RLHFV}.
Abnormal-Aware Rewarding, comprising Relevance Reward, Abnormal Localization Reward and Vision Relevance Reward, aims to improve the model's capability to capture abnormal areas.

In the experiments, we evaluate UMed-LVLM and other methods on MAU dataset, and UMed-LVLM outperforms other competitors in understanding medical images and identifying abnormalities. Moreover, an in-depth analysis revealed that while Med-LVLMs are not yet adept at abnormality detection in medical images, enhancing their abnormality detection capabilities can improve their understanding of medical images. Furthermore, our findings suggest that large models possess out-of-distribution (OOD) generalization capabilities, indicating the potential for more robust disease recognition through iterative improvements in large models, even if they are only exposed to diverse medical images of limited disease types \cite{DBLP:conf/miccai/ZhangDR21}. In addition, we analyze the cross-modal capabilities and generalization potential of Med-LVLMs, and results show that augmenting models in varied modalities enhances their performance in modality with limited dataset availability.

The main contributions of this work are below:
\begin{itemize}
\item We propose a novel UMed-LVLM for medical diagnosis with unveiling abnormality by enhancing its visual localization capability.
\item We introduce Abnormal-Aware Instruction Tuning and Abnormal-Aware Rewarding strategy to train the Med-LVLM, aiming to enhance the model's focus on abnormal areas when generating responses. 
\item We design a prompt method to create the MAU dataset for UMed-LVLM training. The dataset comprises medical images and diagnoses with abnormality annotations.
\item Experimental results show that UMed-LVLM outperforms existing Med-LVLMs in identifying and understanding medical abnormalities. We conduct an in-depth analysis of model training and generalization capabilities, underscoring the potential of incorporating medical abnormalities to enhance Med-LVLM.
\end{itemize}

\section{Related Work}
Recent advancements in LVLMs have significantly improved visual comprehension and contextual language understanding \citep{alayrac2022flamingo,chen2022pali,zhou2024visual}. Notable models like CLIP \citep{pmlr-v139-radford21a} and BLIP-2 \citep{BLIP-2} have achieved impressive results in vision-language tasks by leveraging pre-trained image-text pairs. The introduction of GPT-4 \citep{openai2023gpt4} has further propelled this field, with models like LLaVA \citep{liu2023visual} and its improved version \citep{liu2023improved} demonstrating exceptional capabilities in multimodal tasks. In the medical domain, models such as LLaVA-Med \citep{LLaVA-Med}, Visual Med-Alpaca \citep{shu2023visual}, OphGLM \citep{gao2023ophglm}, and XrayGPT \citep{thawkar2023xraygpt} have specialized in interpreting medical images and providing comprehensive assistance. Observations in \citep{wu2023can} indicated that although GPT-4V performs well in differentiating medical image modalities and anatomy, it still has difficulties in disease diagnosis and generating comprehensive reports. Additionally, Reinforcement Learning (RL) has been applied to LLMs to enhance their performance and flexibility \citep{DBLP:books/lib/SuttonB98, DBLP:journals/nature/MnihKSRVBGRFOPB15, DBLP:conf/icml/SchulmanLAJM15, DBLP:conf/icml/MnihBMGLHSK16, DBLP:journals/corr/SchulmanWDRK17}.  
The full version can be found in Appendix~\ref{app:related}.

\begin{figure}[!t]
  \centering
  \includegraphics[width=\linewidth]{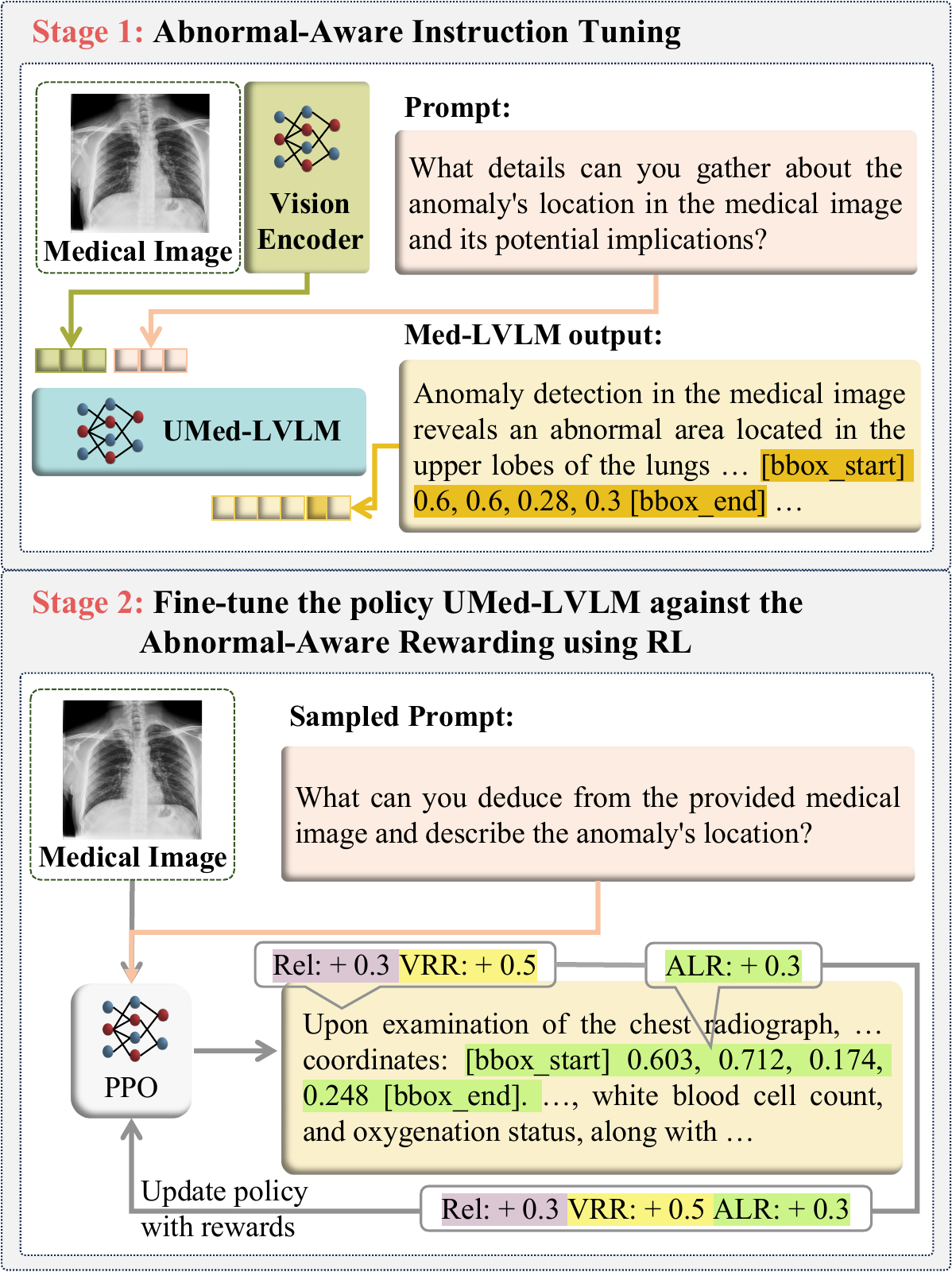}
  \caption{\small The two-stage training for our $\Ours$. ``Rel'', ``ALR'' and ``VRR'' denote ``Relevance Reward'', ``Abnormal Localization Reward'' and ``Vision Relevance Reward'', respectively, as detailed in Sec.~\ref{sec:AAR}.}
  \label{fig:pipeline}
\end{figure}
\section{Methodology}
UMed-LVLM is trained via a two-stage training method, i.e., Abnormal-Aware Instruction Tuning and Abnormal-Aware Rewarding. 
UMed-LVLM is continually trained on MedVInt~\cite{DBLP:journals/corr/abs-2305-10415}. 
Abnormal-Aware Rewarding comprises LLM relevance rewarding, Abnormal Localization Rewarding, and Vision Relevance Rewarding.

\subsection{Abnormal-Aware Instruction Tuning}
We introduce an abnormal-aware instruction tuning approach to improve the model's ability to understand medical abnormal regions and generate corresponding diagnoses, building upon previous work in instruction tuning \cite{DBLP:conf/iclr/WeiBZGYLDDL22, liu2023visual}.  
Given a medical image $x$ and a user query $q$, model generates a response $a$ token-by-token. The response includes the diagnosis and a description of the abnormal region, i.e.,
\begin{align}
    \vp(a|x, q; \theta) = \prod_{t=1}^T \vp(a_t|a_{<t}, x, q; \theta),
\end{align}
where $\theta$ represents the model parameters, and $T$ is the length of the response.

During training, we provide the model with medical images, user queries, and responses including diagnoses and descriptions of abnormal regions. Model are optimized by cross-entropy loss, i.e.,
\begin{align}
  \gL_{it} = -\sum_{i=1}^{T} \log \vp_{i}.
\end{align}
where $\vp_{i}$ is the probability of the $i$-th token in response $a$.
This approach improves the model's understanding of abnormal regions in medical images and its ability to provide diagnosis outputs with identified abnormal regions. However, it does not directly guide the model to focus on abnormal regions in medical images.

\subsection{Abnormal-Aware Rewarding}\label{sec:AAR}
To address the challenge of effectively identifying and describing abnormalities within medical images, we propose an Abnormal-Aware Rewarding (AAR) strategy for $\Ours$ training. This reinforcement learning (RL) training strategy comprises three rewarding strategies, i.e., Relevance Rewarding, Abnormal Localization Rewarding and Vision Relevance Rewarding, designed to optimize the Med-LVLMs based on abnormalities. In contrast to \citet{InstructGPT} optimize LLMs following user instructions by the Proximal Policy Optimization (PPO \cite{DBLP:journals/corr/SchulmanWDRK17}), AAR optimizes the Med-LVLMs by a more directed learning towards the accurate medical diagnosis with abnormality recognition.

\paragraph{Relevance Rewarding.}
The relevance rewarding framework is fundamentally structured around three pivotal components: the policy network, the value network and the LLM reward model. Both networks play a crucial role in guiding the training process, with the policy network ($\pi$) generating actions (responses) based on the given state $s_t$, which encapsulates the medical image $x$ and user query $q$. The policy network's output is mathematically expressed as:
\begin{align}
  \pi(a_t|s_t; \theta) = \prod_{t=1}^T \pi(a_t|s_t, a_{<t}; \theta),
\end{align}
where $\theta$ denotes the parameters of policy network.

Simultaneously, the value network ($V$) is tasked with estimating the expected return from state $s_t$, offering a benchmark for calculating advantage function vital for optimizing the policy network:
\begin{align}
  V(s_t; \phi) = \mathbb{E}\left[r_t \mid s_t, \pi \right],
\end{align}
where $\phi$ represents the parameters of the value network, and $r_t$ denotes the immediate reward associated with the current state $s_t$. 

The LLM relevance rewarding framework incorporates rewards from the policy network, the value network, and the LLM relevance reward model. The total reward, $r_t^{\pi,V,LLM}$, measures the improvement of the chosen action over the baseline provided by the value network and includes the relevance reward from the LLM model:
\begin{align}
r_t^{\pi,V,LLM} = A(s_t, a_t; \theta, \phi) + r_t^{LLM},
\end{align}
where $A(s_t, a_t; \theta, \phi) = Q(s_t, a_t; \theta) - V(s_t; \phi)$ is the advantage function, and $r_t^{LLM}$ represents the relevance reward provided by the LLM model. The Q-function \( Q(s_t, a_t; \theta) \) is updated using the Bellman equation:
\begin{align}
\Delta &Q(s_t, a_t) = \notag \\
&\alpha \left[ r_t + \gamma \max_{a'} Q(s_{t+1}, a')-Q(s_t, a_t) \right],\\
&Q(s_t, a_t) \leftarrow Q(s_t, a_t) + \Delta Q(s_t, a_t),
\end{align}
where $\alpha$ is the learning rate, $r_t$ is the immediate reward, $\gamma$ is the discount factor, and \( \max_{a'} Q(s_{t+1}, a') \) is the maximum expected future reward for the next state \( s_{t+1} \). 

\paragraph{Abnormal Localization Reward.}
The Abnormal Localization Reward (ALR), denoted as $r_t^{loc}$, quantifies the accuracy of predicted bounding boxes by calculating the Intersection over Union (IoU) between the predicted bounding box ($\text{Pred-BBox}$) and the ground truth bounding box ($\text{GT-BBox}$).  Specifically, $r_t^{loc}$ is defined as:
\begin{align}
  r_t^{loc} = \frac{\text{Overlap}(\text{Pred-BBox}, \text{GT-BBox})}{\text{Union}(\text{Pred-BBox}, \text{GT-BBox})}. \label{equ:iou}
\end{align}
This reward directly encourages the model to accurately localize the regions of abnormality within the medical image. A higher IoU score indicates a greater overlap between the predicted and ground truth bounding boxes, signifying a more precise localization of the abnormal area.  

\begin{figure*}[t]
  \centering
  \includegraphics[width=\linewidth]{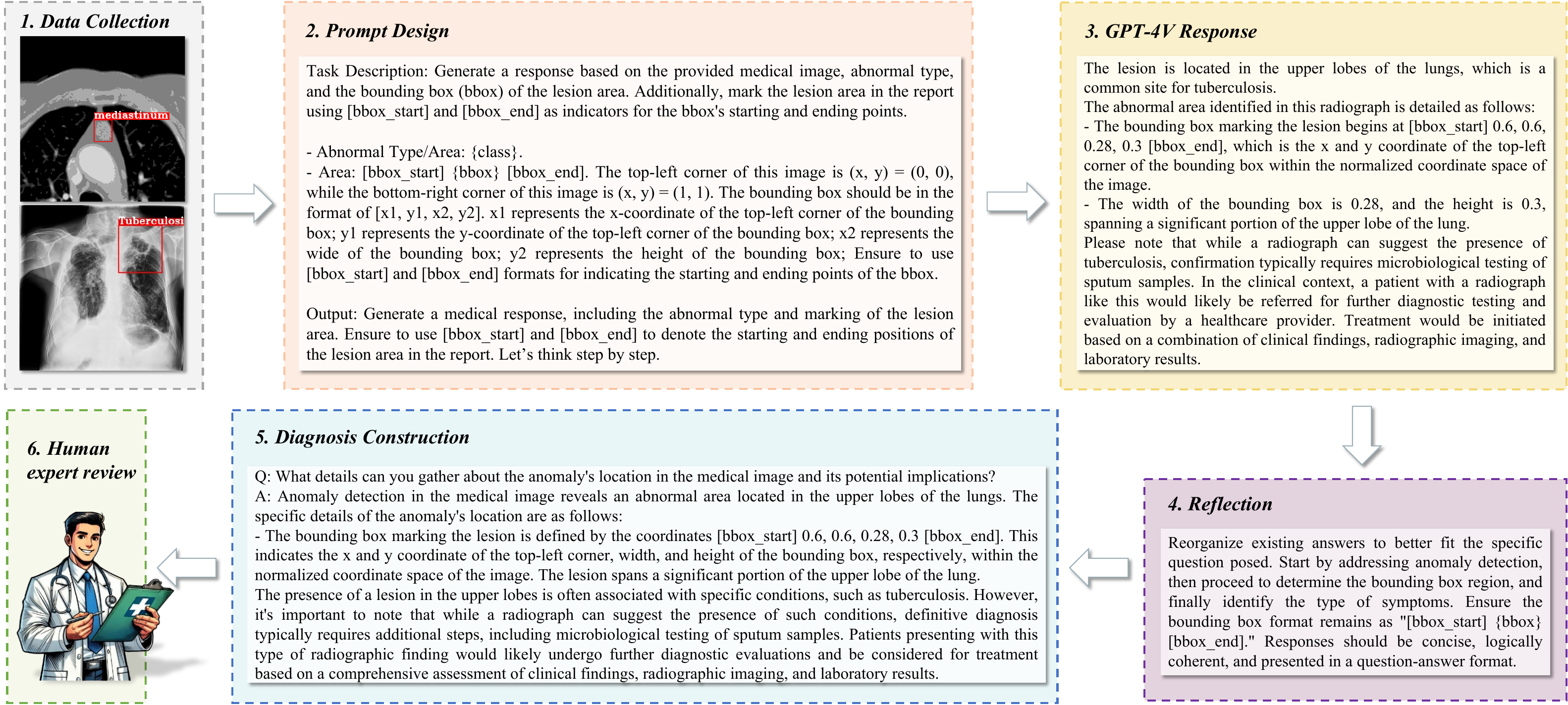}
  \caption{\small Pipeline overview for constructing the medical abnormalities unveiling (MAU) dataset. The process involves data collection, prompt design, GPT-4V response generation, reflection for outputs, diagnosis construction, and human expert review.}
  \label{fig:prompt}
\end{figure*}

\paragraph{Vision Relevance Reward.}
The Vision Relevance Reward (VRR), $r_t^{att}$, is computed by aggregating attention weights between abnormal category tokens and image patches identified as abnormal regions. Specifically, within a transformer framework, the VRR quantifies the model's focus on crucial visual areas by analyzing the attention scores allocated to tokens corresponding to abnormal categories and their association with image patches indicative of abnormalities. This mechanism enhances diagnostic accuracy by reinforcing the model's concentration on critical visual information. The VRR is calculated as:
\begin{align}
  r_t^{att} = \sum_{i \in N} \sum_{j \in \bar{N}} \frac{\exp\left(\frac{Q_i \cdot K_j^\top}{\sqrt{d_k}}\right)}{\sum_{k \in \bar{N}} \exp\left(\frac{Q_i \cdot K_k^\top}{\sqrt{d_k}}\right)},
\end{align}
where $N$ denotes the set of tokens associated with identified textual abnormalities, and $\bar{N}$ represents the set of image patches characterized as abnormal regions.  $Q_i$ and $K_j$ are the query vector for token $i$ and the key vector for image patch $j$, respectively; $d_k$ is the dimensionality of the key vectors. This approach leverages the transformer's attention mechanism to quantify the model's focus on critical areas. By emphasizing the importance of abnormality-related tokens and their corresponding attention weights over abnormal image patches, the model is incentivized to attend to visual information indicative of medical abnormalities, ultimately improving diagnostic performance.

\paragraph{Reward Normalization and Aggregation.}
To achieve equilibrium between the ALR ($r_t^{loc}$) and the VRR ($r_t^{att}$), we individually normalize these rewards for responses generated by the same query. This normalization ensures that each reward type contributes equally to the final reward calculation. The combined reward for each response is computed as follows:
\begin{align}
  r_t = r_t^{\pi,V,LLM} + \frac{r_t^{loc}}{\max(r_t^{loc})} + \frac{r_t^{att}}{\max(r_t^{att})},
\end{align}
where $\max(r_t^{loc})$ and $\max(r_t^{att})$ are the maximum values of the abnormal-aware localization and attention reward, respectively, for all responses to a particular query.

\paragraph{Optimization Process.}
As an improved version of the PPO, our policy network is refined by maximizing an objective function that incorporates the combined reward $r_t$ into the PPO. Specifically, we aim to maximize an objective function that combines the clipped surrogate objective with an entropy bonus to encourage exploration:
\begin{align}
\notag \gL^{\text{CLIP}+\text{ENT}}(\theta) =& \hat{\mathbb{E}}\Big[\gL^{\text{CLIP}}(\theta) + c_{1} r_t - c_{2} \gL^{VF}(\phi) \\
&+ c_{3} S[\pi(\cdot|s_t)]\Big],\label{equ:reward}
\end{align}
where $\gL^{\text{CLIP}}(\theta)$ is the clipped part of the PPO objective, $r_t$ is the combined normalized reward as defined previously, $\gL^{VF}(\phi)$ is the value function loss, $S[\pi(\cdot|s_t)]$ denotes the entropy of the policy for state $s_t$, and $c_{1}$, $c_{2}$, $c_{3}$ are coefficients balancing the contribution of each term.

\section{MAU Dataset}
To enhance the abnormality unveiling capabilities of the Med-LVLM, we construct the MAU dataset for $\Ours$ training. Firstly, we collect medical image datasets with abnormal annotations. Then, we design a Prompt Method to construct the MAU dataset, a medical diagnosis dataset with abnormal annotations, by GPT-4V \citep{2023GPT4VisionSC}. 

\paragraph{Collecting Medical Image Datasets with Abnormal Areas.}
We collect a medical image dataset annotated with abnormal areas for training the Med-LVLM. This dataset encompasses five distinct sub-datasets, namely DeepLesion \cite{DBLP:journals/corr/abs-1710-01766}, KidneyStone \citep{tez_roi_aug_dataset}, NIH \cite{DBLP:conf/cvpr/WangPLLBS17}, TBX11K \cite{DBLP:conf/cvpr/LiuWBWC20}, and KVASIR \cite{DBLP:conf/mmsys/PogorelovRGELJS17}, each originating from diverse sources.
DeepLesion includes 32,120 axial CT slices with eight types of abnormalities. The KidneyStone dataset contains 1,300 renal CT scans of various kidney stones in different sizes, shapes, and locations. The NIH dataset has 112,120 chest X-ray images across 14 pathological categories. The TBX11K dataset consists of 11,200 chest X-ray images for tuberculosis detection. The KVASIR dataset provides 8,000 endoscopic images of the gastrointestinal tract, with eight types of abnormalities.
These sub-datasets span various medical imaging modalities, including X-rays, CT scans, and gross pathology. We utilized a portion of these datasets, and Appendix~\ref{app:dataset} summarizes the specific details of the segments used from each sub-dataset. Each image is annotated with the type of abnormality present and includes bounding box information for identified abnormal areas.

\paragraph{Medical Abnormal Unveiling Dataset Construction.}
To construct the Medical Abnormal Unveiling (MAU) Dataset, we design a Prompt Method that utilizes the GPT-4V model to generate diagnosis annotations with medical abnormal areas. The pipeline, as shown in Figure \ref{fig:prompt}, comprises two stages: diagnosing abnormalities in medical images and reflecting on the previous diagnosis. Firstly, we integrate collected medical images, corresponding abnormality categories, and the locations of abnormal areas into our designed prompt. This prompt is then passed into GPT-4V to generate diagnosis responses based on the given abnormality categories and areas. 
To build a diagnosis with step-by-step thoughts, a reflection prompt is designed to reorganize these diagnosis responses, starting from abnormality detection, identifying bounding box regions, and finally recognizing abnormality categories. 
The processed responses, the medical images, and queries form medical image diagnosis samples. 
Through this Prompt Method, we use GPT-4V to generate an MAU dataset (examples in Appendix~\ref{app:dataset}). Our plug-and-play prompt method easily integrates with other medical datasets and is disease-agnostic.

\paragraph{Review by Human Experts.}
To ensure the reliability of the MAU dataset, all generated data have been reviewed and filtered by human expert reviewers (i.e., three doctoral students specializing in medicine). During review, only 13 samples were found to have errors, which were corrected manually by reviewers. 
As shown in Table~\ref{tab:main}, using GPT-4V for direct diagnosis without additional information yields limited performance. However, providing the model with the location of the abnormal area leads to a significant performance improvement. Crucially, in our dataset construction, we provide GPT-4V with both the abnormal area and the ground-truth diagnosis. GPT-4V's role is not to perform the diagnosis itself, but rather to generate an explanation (the ``diagnosis'' text) of the reasoning process, given the known diagnosis and the location of the abnormality. This approach leverages the model's strong language generation capabilities while mitigating its limitations in direct medical image interpretation, resulting in a dataset of high quality and reliability.

\section{Experiments}
\subsection{Experimental Setups}
We evaluate our method on five benchmarks: test set of our MAU, VQA-RAD \cite{VQA-RAD} and SLAKE \cite{Slake}, PMC-VQA-test~\cite{DBLP:journals/corr/abs-2305-10415} and MedMNIST \cite{MedMNIST}. 
Our MAU comprises five sub-datasets: DeepLesion, KidneyStone, NIH, TBX11K, and KVASIR. 
We utilized MedVInt \cite{zhang2023pmc} as the initialization for our model. 
UMed-LVLM is trained on the training set of 4,653 examples and then evaluated on 1,164 examples on MAU. Crucially, we ensure no case overlap between the training and test sets, based on the original data annotations, to prevent data leakage. We employ Accuracy (ACC) as the evaluation metric. 
ACC assesses the correctness of diagnosis results, i.e., whether the response text includes the correct category of abnormalities, following \cite{doi:10.1148/radiol.232255} to evaluate the correctness of outputs from LVLMs.
The implementation details and details of the compared LVLMs methods are in Appendix~\ref{app:setting} and Appendix~\ref{app:lvlms}.

\subsection{Results and Discussion}
\begin{table}[t]\small
\centering
\resizebox{\linewidth}{!}{
\setlength{\tabcolsep}{2.5pt}
\begin{tabular}{lccccccc}
\toprule
\bf Method        & \bf DL & \bf KS & \bf KV & \bf NIH  & \bf TBX  & \bf Avg. \\
\midrule
MiniGPT-4     & 0.02         & 0.00          & 0.02   & 0.00 & 0.00 & 0.01    \\
mPLUG-Owl     & 0.05         & 0.00          & 0.01   & 0.00 & 0.00 & 0.01    \\
LLaVA         & 0.20         & 0.00          & 0.04   & 0.00 & 0.00 & 0.05    \\
Qwen-VL       & 0.13         & 0.00          & 0.01   & 0.00 & 0.00 & 0.03    \\\midrule
XrayGPT       & 0.18         & 0.12          & 0.02   & 0.07 & 0.06 & 0.09    \\
LLaVA-Med     & 0.22         & 0.04          & 0.12   & 0.03 & 0.01 & 0.08    \\
Med-Flamingo  & 0.27         & 0.15          & 0.15   & 0.09 & 0.02 & 0.14    \\
MedVInt       & 0.29         & 0.11          & 0.27   & 0.08 & 0.09 & 0.17    \\\hline\midrule
\multicolumn{6}{c}{\textit{Trained on MAU Dataset}}  \\\midrule
MedVInt (SFT)  & 0.42 & 0.93 & 0.93 & 0.28 & 0.78 & 0.67 \\
MedVInt (SFT\&PPO)  & 0.44         & 0.94          & 0.95   & 0.30 & 0.80 & 0.69    \\
\rowcolor{gray!15}UMed-LVLM     & \bf 0.53         & \bf 0.99          & \bf 0.98   & \bf 0.37 & \bf 0.86 & \bf 0.75    \\\hline\midrule
\multicolumn{6}{c}{\textit{Closed-Source Model}}  \\\midrule
GPT-4V        & 0.27         & 0.36          & 0.53   & 0.18 & 0.19 & 0.31    \\
GPT-4V w/ bbox & 0.50         & 0.95         & 0.95   & 0.32 & 0.81 & 0.72    \\
\bottomrule
\end{tabular}}
\caption{\small Comparison on test set.``DL'', ``KS'' and ``KV'' denote ``Deep Lesion'', ``Kidney Stone'' and ``KVASIR'', respectively. 
The comparison models include LVLMs (i.e., MiniGPT-4~\cite{zhu2023minigpt}, mPLUG-Owl~\cite{ye2023mplug}, LLaVA~\cite{liu2023visual}, Qwen-VL~\cite{bai2023qwen}, GPT-4V~\cite{2023GPT4VisionSC}) and Med-LVLMs (i.e., XrayGPT~\cite{thawkar2023xraygpt}, LLaVA-Med~\cite{LLaVA-Med}, Med-Flamingo~\cite{DBLP:conf/ml4h/MoorHWYDLZRR23}, MedVInt~\cite{DBLP:journals/corr/abs-2305-10415}). 
MedVInt (SFT) denotes MedVInt trained with SFT on the MAU dataset, while MedVInt (SFT\&PPO) indicates MedVInt trained with SFT and PPO on the MAU dataset. GPT-4V w/ bbox denotes that GPT-4V was provided with abnormal region location information.}
\label{tab:main}
\end{table}

\begin{table}[t]\small
\centering
\resizebox{\linewidth}{!}{
\setlength{\tabcolsep}{2.2pt}
\begin{tabular}{lcccc}
\toprule
\multirow{2}{*}{\textbf{Method}} & \multicolumn{2}{c}{\bf VQA-RAD} & \multicolumn{2}{c}{\bf SLAKE} \\
\cmidrule(lr){2-3} \cmidrule(lr){4-5}
 & \bf Open & \bf Close & \bf Open & \bf Close \\
\midrule
MEVF-BAN \cite{MEVF-BAN} & 49.2 & 77.2 & 77.8 & 79.8 \\
CPRD-BAN \cite{CPRD-BAN} & 52.5 & 77.9 & 79.5 & 83.4 \\
M3AE \cite{M3AE} & 67.2 & 83.5 & 80.3 & 87.8 \\
PMC-CLIP \cite{PMC-CLIP} & 67.0 & 84.0 & 81.9 & 88.0 \\
MedVInT~\cite{DBLP:journals/corr/abs-2305-10415} & 69.3 & 84.2 & 88.2 & 87.7 \\
\rowcolor{gray!15}UMed-LVLM & \bf 74.9 & \bf 87.6 & \bf 90.4 & \bf 89.5 \\
\bottomrule
\end{tabular}}
\caption{\small Results on VQA-RAD \cite{VQA-RAD} and SLAKE \cite{Slake}.}
\label{tab:vqa}
\end{table}

\begin{table}[t]\small
\centering
\begin{tabular}{lcc}
\toprule
\bf Method & \bf Choice & \bf Blanking \\
\midrule
PMC-CLIP \cite{PMC-CLIP} & 24.7 & - \\
BLIP-2 \cite{BLIP-2} & 24.3 & 21.8 \\
Open-Flamingo \cite{openai2023gpt4} & 26.4 & 26.5 \\
LLAVA-Med \cite{LLaVA-Med} & 34.8 & 29.4 \\
MedVInT~\cite{DBLP:journals/corr/abs-2305-10415} & 39.2 & 35.3 \\
\rowcolor{gray!15}UMed-LVLM & \bf 42.6  & \bf 38.1 \\
\bottomrule
\end{tabular}
\caption{\small Comparison of different methods on PMC-VQA-test~\cite{DBLP:journals/corr/abs-2305-10415} (metric is accuracy).}
\label{tab:PMC-VQA-test}
\end{table}

\begin{table}[t]\small
\centering
\resizebox{\linewidth}{!}{
\setlength{\tabcolsep}{3pt}
\begin{tabular}{lcccccc}
\toprule
\multirow{2}{*}{\textbf{Method}} & \multicolumn{2}{c}{\textbf{Pneumonia}} & \multicolumn{2}{c}{\textbf{Breast}} & \multicolumn{2}{c}{\textbf{Derma}} \\
\cmidrule(lr){2-3} \cmidrule(lr){4-5} \cmidrule(lr){6-7}
& \bf AUC$\uparrow$ & \bf ACC$\uparrow$ & \bf AUC$\uparrow$ & \bf ACC$\uparrow$ & \bf AUC$\uparrow$ & \bf ACC$\uparrow$ \\
\midrule
ResNet50 & 96.2 & 88.4 & 86.6 & 84.2 & 91.2 & 73.1 \\
DWT-CV   & 95.7 & 88.7 & 89.8 & 85.7 & 91.7 & 74.8 \\
SADAE    & 98.3 & 91.8 & 91.5 & 87.8 & 92.7 & 75.9 \\
PMC-CLIP & 99.0 & 95.4 & 94.6 & 91.4 & 93.4 & 79.8 \\
MedVInT  & 98.5 & 94.9 & 93.4 & 90.4 & 93.7 & 80.0 \\
\rowcolor{gray!15}UMed-LVLM & \bf 99.1    & \bf 95.8    & \bf 95.2     & \bf 92.8    & \bf 94.2   & \bf 84.1   \\
\bottomrule
\end{tabular}}
\caption{\small Performance comparison on MedMNIST \cite{MedMNIST}: Pneumonia (chest X-ray), Breast (ultrasound), and Derma (dermatoscopy).  The comparison models include ResNet50 \cite{ResNet50}, DWT-CV \cite{DWT-CV}, SADAE \cite{SADAE}, PMC-CLIP, and MedVInT.}
\label{tab:MNIST}
\end{table}

As shown in Table~\ref{tab:main}, our method outperforms other methods on the MAU. 
By comparing UMed-LVLM with MedVInt (SFT) and MedVInt (SFT\&PPO), the performance gain can be attributed to the MAU dataset and our Abnormal-Aware Rewarding approach.
From the results, UMed-LVLM significantly outperforms GPT-4V. 
Furthermore, providing ground-truth abnormal region locations to GPT-4V (GPT-4V w/ bbox) leads to a significant improvement, emphasizing the importance of abnormal region localization for diagnosis.
Table~\ref{tab:vqa} and Table~\ref{tab:PMC-VQA-test} present results on VQA-RAD, SLAKE, and PMC-VQA-test datasets. 
UMed-LVLM outperforms other methods, showcasing its ability to understand and reason about medical images and related questions.
In Table~\ref{tab:MNIST}, UMed-LVLM outperforms other methods on MedMNIST (Pneumonia, Breast, Derma).  Remarkably, despite continual training without Ultrasound and Dermatoscopy data, UMed-LVLM generalizes well to these modalities, demonstrating its effectiveness and generalization.
The case study can be found in Appendix~\ref{app:case}.

\subsection{Ablation Study}
To verify the efficacy of components in AAR, we conduct experiments by progressively removing each component from UMed-LVLM to observe the impact on performance. 
As shown in Table~\ref{tab:ablation}, we compare the performance of UMed-LVLM and its three variants, i.e., ``w/o ALR'', ``w/o VRR'', and ``w/o AAR''. The results show a performance decline as components are removed, demonstrating that each component plays a critical role in enhancing diagnosis accuracy for medical images. 
The VRR, by aligning abnormal identification and attention regions, ensures that the model does not overlook subtle but critical abnormalities in the images. 
The performance degradation observed when either component is removed substantiates the hypothesis that a dual-reward system, which addresses abnormality recognition and localization, is beneficial in diagnosis.
\begin{table}[t]\small
\centering
\setlength{\tabcolsep}{4.7pt}
\begin{tabular}{lcccccc}
\toprule
\textbf{Method} & \bf DL & \bf KS & \bf KV & \bf NIH  & \bf TBX  & \bf Avg. \\
\midrule
\rowcolor{gray!15}UMed-LVLM     & \bf 0.53         & \bf 0.99          & \bf 0.98   & \bf 0.37 & \bf 0.86 & \bf 0.75    \\\midrule
w/o VRR & 0.49 & 0.97 & 0.95 & 0.30 & 0.82 & 0.71 \\
w/o ALR & 0.48 & 0.96 & 0.96 & 0.35 & 0.83 & 0.72 \\
w/o AAR & 0.42 & 0.93 & 0.93 & 0.28 & 0.78 & 0.67 \\
\bottomrule
\end{tabular}
\caption{\small Ablation study verifies the effectiveness of components in the Abnormal-Aware Rewarding (AAR), including two components: Abnormal Localization Reward (ALR) and Abnormal-Vision Relevance Reward  (VRR).  ``w/o AAR'' denotes UMed-LVLM only trained on Abnormal-Aware Instruction Tuning.}
\label{tab:ablation}
\end{table}

\subsection{Analysis}
\paragraph{Impact of Abnormal Localization.}
\begin{figure}[t]
  \centering
  \includegraphics[width=0.95\linewidth]{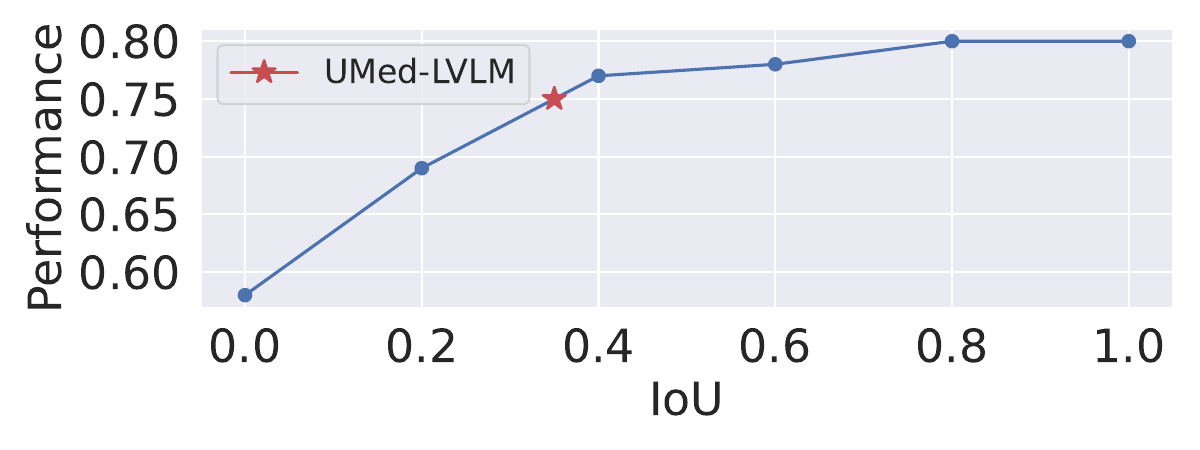}
  \caption{\small Performance of $\Ours$ with different IoUs.}
  \label{fig:IoU}
\end{figure}
To verify the impact of localization ability during the reasoning process on Med-LVLM, we injected bounding boxes (bbox) corresponding to different IoU scores during model inference for observation. The experimental results, as shown in Figure~\ref{fig:IoU}, indicate that enhancing localization ability can influence the diagnostic results. As anomaly localization becomes more accurate, the model's diagnostic performance improves. However, when IoU exceeds 0.6, the improvement in the model's diagnostic performance begins to plateau. This suggests that Med-LVLM does not require extremely high localization accuracy to leverage the reasoning gains from localization; a certain level of localization ability is sufficient to enhance diagnostic reasoning. The red dot in the figure represents the performance achieved by our $\Ours$. From the trend, it is evident that the localization ability of our method effectively enhances model's diagnostic capability.

\paragraph{Impact of Instruction Tuning Epoch.}
\begin{figure}[!t]
  \centering
  \includegraphics[width=0.49\linewidth]{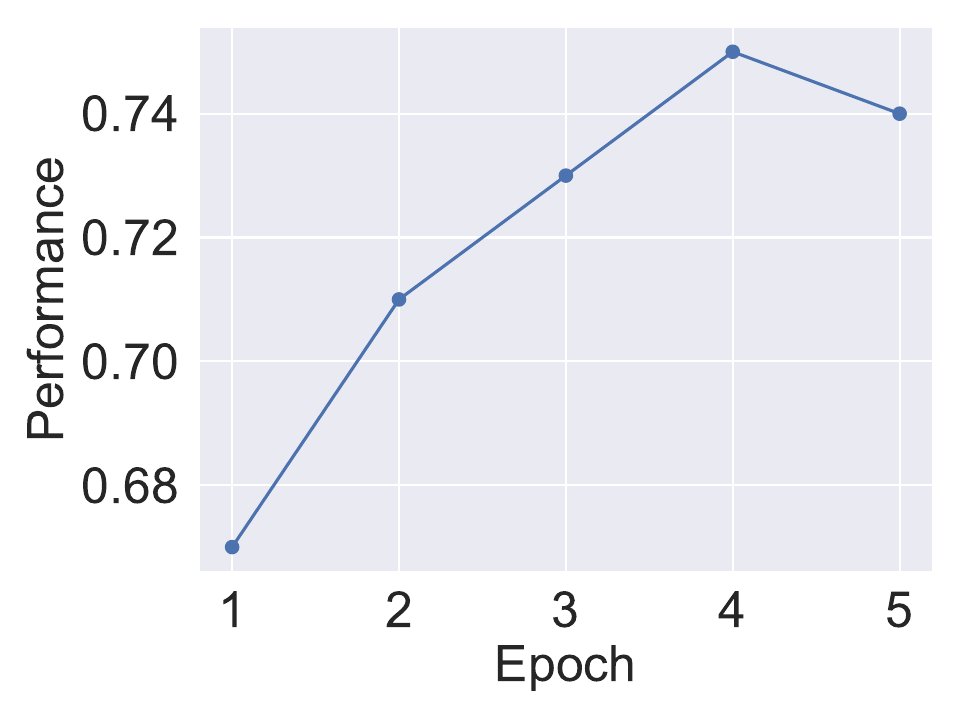}
  \includegraphics[width=0.49\linewidth]{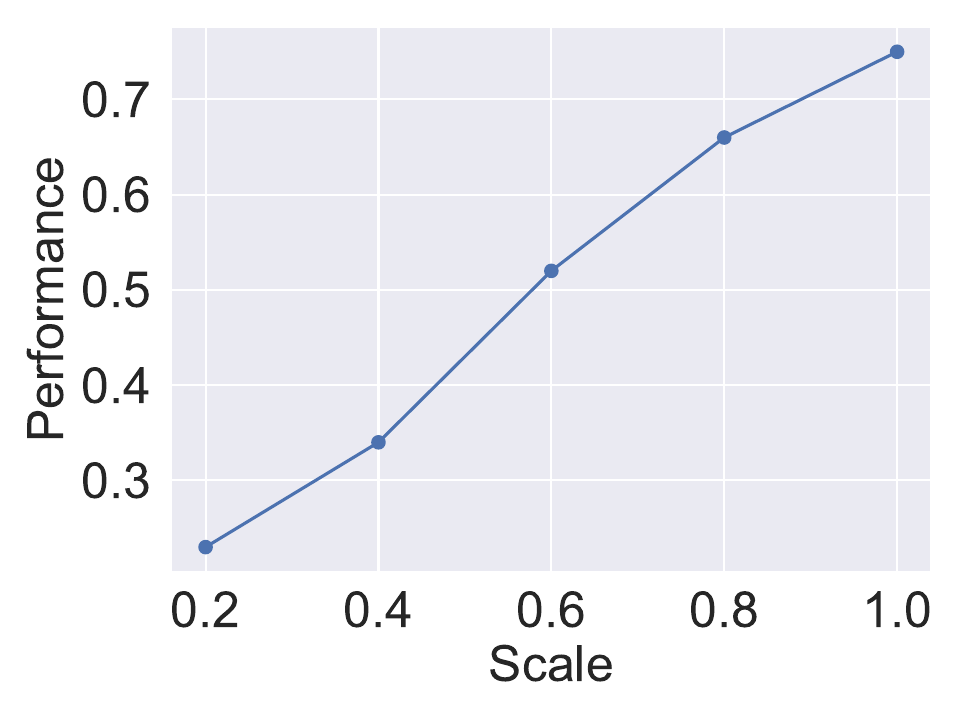}
  \caption{\small Performance of Abnormal-Aware Instruction Tuning across different epochs (Left)  and data scales(Right).}
  \label{fig:epoch}
  \label{fig:scale}
\end{figure}
As shown in Figure~\ref{fig:epoch}(Left), we investigate the effects of employing Instruction Tuning over different epochs on the performance of our model. The performance exhibits an upward trajectory until epoch 4, achieving a peak of approximately 0.75, which suggests that the model becomes progressively more adept at executing tasks as per the user queries and medical image inputs. 

\paragraph{Impact of Training Set Scale for Instruction Tuning.}
As shown in Figure~\ref{fig:scale}(Right), the performance of our model exhibits a positive correlation with the scale of the training data. We can observe a substantial increment as the data scale progresses from 20\% to 100\%. This trend illustrates that the model benefits from a larger volume of data, refining its ability to understand abnormalities and execute instructions more effectively. 
This improvement underscores the model's evolving proficiency in identifying the exact spatial extents of abnormalities, which is crucial for subsequent diagnosis tasks. The incremental nature of this trend indicates that the model's spatial comprehension capabilities can be significantly honed with more extensive training data. These results collectively highlight the critical role that data scale plays in the efficacy of instruction tuning for Med-LVLM. 

\begin{table}[t]\small
\centering
\begin{tabular}{lccc}
\toprule
\bf Method    & \bf Abdomen & \bf Lung & \bf Pelvis \\
\midrule
\rowcolor{gray!15}UMed-LVLM & \bf 0.35    & \bf 0.39 & \bf 0.27 \\
MedVInt (SFT)   & 0.16    & 0.15 & 0.16 \\
MedVInt (SFT\&PPO)   & 0.15    & 0.12 & 0.15 \\
MedVInt   & 0.05    & 0.04 & 0.07 \\
\bottomrule
\end{tabular}
\caption{\small Model generalization to categories excluded from training set. MedVInt (SFT) and MedVInt (SFT\&PPO) can be found in Table~\ref{tab:main}.}
\label{tab:ood}
\end{table}

\paragraph{Generalization Capability for Untrained Category.}
This setting aims to evaluate the generalization capability of the UMed-LVLM on medical categories not present in the training data. Specifically, we removed data in three categories (i.e., ``Abdomen'', ``Lung'', and ``Pelvis'' on ``DeepLesion'') on the training set. Then we evaluated the trained model on these categories to observe its ability to handle untrained categories. The results, shown in Table~\ref{tab:ood}, present the performance in each category. The performance demonstrates that our method exhibits a degree of generalization capability on untrained categories of medical images. 
Although the performance on untrained categories is somewhat reduced compared to training on these categories, these findings indicate that the model can generalize knowledge learned during training to new categories not present in the training data. 
In contrast, MedVInt shows a slight ability to generalize to these untrained categories with scores close to zero across all categories. It demonstrates the effectiveness of abnormal-aware learning.

\paragraph{Generalization Capability of Abnormality Unveiling.}
\begin{table}[t]\small
\centering
\setlength{\tabcolsep}{8pt}
\begin{tabular}{lcc}
\toprule
\bf Method    & \bf TBX11K & \bf Deep Lesion \\
\midrule
\rowcolor{gray!15}UMed-LVLM & \bf 0.57   & \bf 0.42 \\
MedVInt (SFT)   & 0.29   & 0.28 \\
MedVInt (SFT\&PPO)   & 0.30   & 0.22 \\
MedVInt   & 0.10   & 0.08 \\
\bottomrule
\end{tabular}
\caption{\small Performance of UMed-LVLM w/o training on TBX11K and DeepLesion datasets, respectively. Performance is evaluated on the excluded datasets.}
\label{tab:crossmodal}
\end{table}
To assess the robustness and flexibility of UMed-LVLM and MedVInt, we respectively removed the TBX11K and DeepLesion datasets from the training data to verify the models' generalization capabilities.
We evaluated the models on the excluded datasets to determine their ability to generalize across different medical datasets. 
As shown in Table \ref{tab:crossmodal}, UMed-LVLM demonstrated significantly better generalization than other methods.
The performance disparity shows our model's effectiveness in adapting to varied medical scenarios.

\paragraph{Generalization Capability for Cross-Modality.}
To verify the generalization capabilities of UMed-LVLM, we train it on single-modality medical images and evaluate it across different modalities. As shown in Table~\ref{tab:general}, our approach demonstrates better performance in cross-modal generalization compared to other methods. The performance of our model is attributed to our abnormal-aware training method that enhances the model's capability to adapt to various medical images. The Abnormal-Aware Instruction Tuning and AAR boost the model's capability to localize abnormalities and improve diagnostic accuracy in medical images.
This is particularly important in scenarios where the modalities differ substantially (e.g., CT vs. Gross Pathology). 
\begin{table}[t]\small
\centering
\setlength{\tabcolsep}{5pt}
\begin{tabular}{lccc}
\toprule
\bf Method    & \textbf{C -\textgreater GX} & \textbf{G -\textgreater CX} & \textbf{X -\textgreater CG} \\
\midrule
\rowcolor{gray!15}UMed-LVLM & \bf 0.27  & \bf 0.31  & \bf 0.22 \\
MedVInt (SFT)   & 0.12  & 0.08  & 0.13 \\
MedVInt (SFT\&PPO)   & 0.09  & 0.08  & 0.11 \\
MedVInt   & 0.05  & 0.04  & 0.07 \\
\bottomrule
\end{tabular}
\caption{\small Cross-modal generalization capability models trained on single modality medical images. ``C'', ``G'', and ``X'' represent CT, Gross Pathology, and X-ray, respectively.}
\label{tab:general}
\end{table}

\section{Conclusion}
This study introduces UMed-LVLM, a novel Med-LVLM designed to enhance medical diagnosis by the visual localization of abnormalities in medical images. Through a specialized training process involving the collection of a Medical Abnormalities Unveiling dataset and the implementation of Abnormal-Aware Instruction Tuning and Abnormal-Aware Rewarding. Results show UMed-LVLM surpasses existing Med-LVLMs in accurately detecting and interpreting medical anomalies. The AAR incorporates innovative reward mechanisms that sharpen the model's focus on abnormal areas, thereby improving diagnostic reliability and interpretability. Furthermore, the in-depth analysis demonstrates the generalization capability of UMed-LVLM.

\section*{Limitations}\label{app:limitation}
Our study's limitation is the limited computational resources. Recent advancements have led to the development of increasingly large LVLMs, which require substantial computational power to train and deploy effectively. As a result of these computational limitations, we were unable to apply our methodologies to the largest and most complex open-source models currently available. This limits our ability to expand and validate our approach on larger-scale LVLMs.

\bibliography{custom}
\clearpage
\appendix
\section{Related Work}\label{app:related}
\subsection{Large Vision-Language Models}
Recently, there have been remarkable advancements within the domain of LVLMs \citep{alayrac2022flamingo,chen2022pali,zhou2024rethinking,zhou2024less}. These models effectively bridge the gap between visual comprehension and contextual language understanding, presenting a robust solution for reconciling disparities between visual and textual data, and thereby, enhancing their capacity to address various vision-language tasks.
By pre-training on image-text pairs, CLIP \citep{pmlr-v139-radford21a} achieved zero-shot transfer to diverse computer vision tasks without requiring task-specific training. In contrast, BLIP-2 \citep{BLIP-2} employed frozen pre-trained image encoders and LLMs to bridge the modality gap, achieving effective representation and generative learning. The introduction of GPT-4 \citep{openai2023gpt4} ushered in a new era of LVLMs \citep{zhu2023minigpt,ye2023mplug,bai2023qwen}. LLaVA \citep{liu2023visual} leveraged a language-only GPT-4 to generate high-quality multimodal instruction data effectively. Notably, LLaVA demonstrated exceptional chat capabilities, even when provided with novel images and instructions. The improved version LLaVA \citep{liu2023improved} elevated its performance by incorporating academic-task-oriented VQA datasets \citep{Marino_2019_CVPR,8978122,sidorov2020textcaps,schwenk2022okvqa} and simple response formatting prompts, thus establishing stronger baselines. It is worth mentioning that the inclusion of region-level VQA datasets \citep{kazemzadeh-etal-2014,10.1007/s11263-016-0981-7} could significantly enhance the model's ability to localize fine-grained visual details precisely.

\subsection{Medical Large Vision-Language Models}
With the rapid development of LVLMs, their applications in the medical domain have also increased significantly.
LLaVA-Med \citep{LLaVA-Med} extended the capabilities of LLaVA \citep{liu2023visual} to the medical domain with excellent multimodal dialogue capabilities. Unlike Visual Med-Alpaca \citep{shu2023visual}, which connected image captioning models with an LLM and employed a classifier for model assignment, LLaVA-Med is an end-to-end model specifically designed for the medical field. In specific medical domains, LVLMs also play a significant role. In ophthalmology, OphGLM \citep{gao2023ophglm} combined visual and language capabilities to provide comprehensive ophthalmic assistance. To solve open-ended questions about chest radiographs, XrayGPT \citep{thawkar2023xraygpt} effectively aligned the medical visual encoder \citep{wang2022medclip} with a fine-tuned large language model \citep{chiang2023vicuna} by employing a straightforward linear transformation. The scarcity of non-English language models prompted the introduction of Qilin-Med-VL \citep{liu2023qilin} as the Chinese large vision-language model that combines a pre-trained Vision Transformer \citep{dosovitskiy2021an} with a foundational Large Language Model. Observations in \citep{wu2023can} indicated that although GPT-4V performs well in differentiating medical image modalities and anatomy, it still has difficulties in disease diagnosis and generating comprehensive reports. More specifically, GPT-4V demonstrates inadequate performance in accurately identifying the structures or abnormalities in medical images. In addition, several studies have focused on aligning models with clinical preferences, as highlighted by \citet{Clinician} who emphasize the importance of clinician preference alignment in VLM fine-tuning.  MMedPO \citep{MMedPO} proposes a clinical-aware multimodal preference optimization method to improve the model's understanding of clinical data.  Addressing the critical issue of factual accuracy, \citet{RULE} introduce a reliable multimodal retrieval-augmented generation (RAG) approach.  To facilitate the development of medical VLMs, \citet{MedTrinity} present MedTrinity-25M, a large-scale multimodal dataset with multi-granular annotations.  Furthermore, CARES \citep{CARES} offers a comprehensive benchmark for evaluating the trustworthiness of medical VLMs.  In contrast to these works, our approach focuses on leveraging abnormal regions to guide model optimization.

\subsection{Reinforcement Learning for Large Language Models}
Reinforcement Learning (RL) can be defined as a training paradigm that could learn from interactions with environments \cite{DBLP:books/lib/SuttonB98}.
Following the inception of RL, a multitude of approaches have been proposed, some of the more widely used are Q-learning \cite{DBLP:journals/nature/MnihKSRVBGRFOPB15}, Trust Region Policy Optimization (TRPO) \cite{DBLP:conf/icml/SchulmanLAJM15}, Asynchronous Advantage Actor-critic (A3C) \cite{DBLP:conf/icml/MnihBMGLHSK16} and Proximal Policy Optimization (PPO) \cite{DBLP:journals/corr/SchulmanWDRK17}.
Different from the distribution modeling objective of supervised and unsupervised learning, RL is more flexible in terms of reward functions, which facilitates the application of RL methods to generative AI models \cite{DBLP:journals/corr/abs-2308-14328}. In the domain of LLMs, RL can be a vital technique for their development to unlock significant potential. As a successful application of RL combined with LLM, WebGPT \cite{DBLP:journals/corr/abs-2112-09332} optimized the reward function through RL and rejection sampling with the aim of improving its behavior. Another illustrative example of employing RL to LLM is InstructGPT\cite{InstructGPT}, which used methodology of both reinforcement learning with human feedback (RLHF \cite{DBLP:journals/corr/abs-2009-01325,DBLP:conf/nips/ChristianoLBMLA17}) and PPO to proficiently fine-tune GPT-3 \cite{DBLP:conf/nips/BrownMRSKDNSSAA20}, thereby empowering it to follow a diverse range of written instructions. It is worth noting that, in addition to the realm of general LLMs, there are already some examples of medical LLMs that make use of RL. Some brilliant works about medical LLMs like ClinicalGPT \cite{DBLP:journals/corr/abs-2306-09968} and Zhongjing \cite{DBLP:conf/aaai/YangZZZXJZ24} also utilized RL to reduce the bias of LLM and improve performance. While RL has been extensively harnessed in LLMs and medical LLMs, its potential for enhancing capacity of identifying and describing abnormalities within medical images remains to be explored.

\section{MAU Dataset}\label{app:dataset}

\begin{figure*}[!t]
\centering
\includegraphics[width=\linewidth]{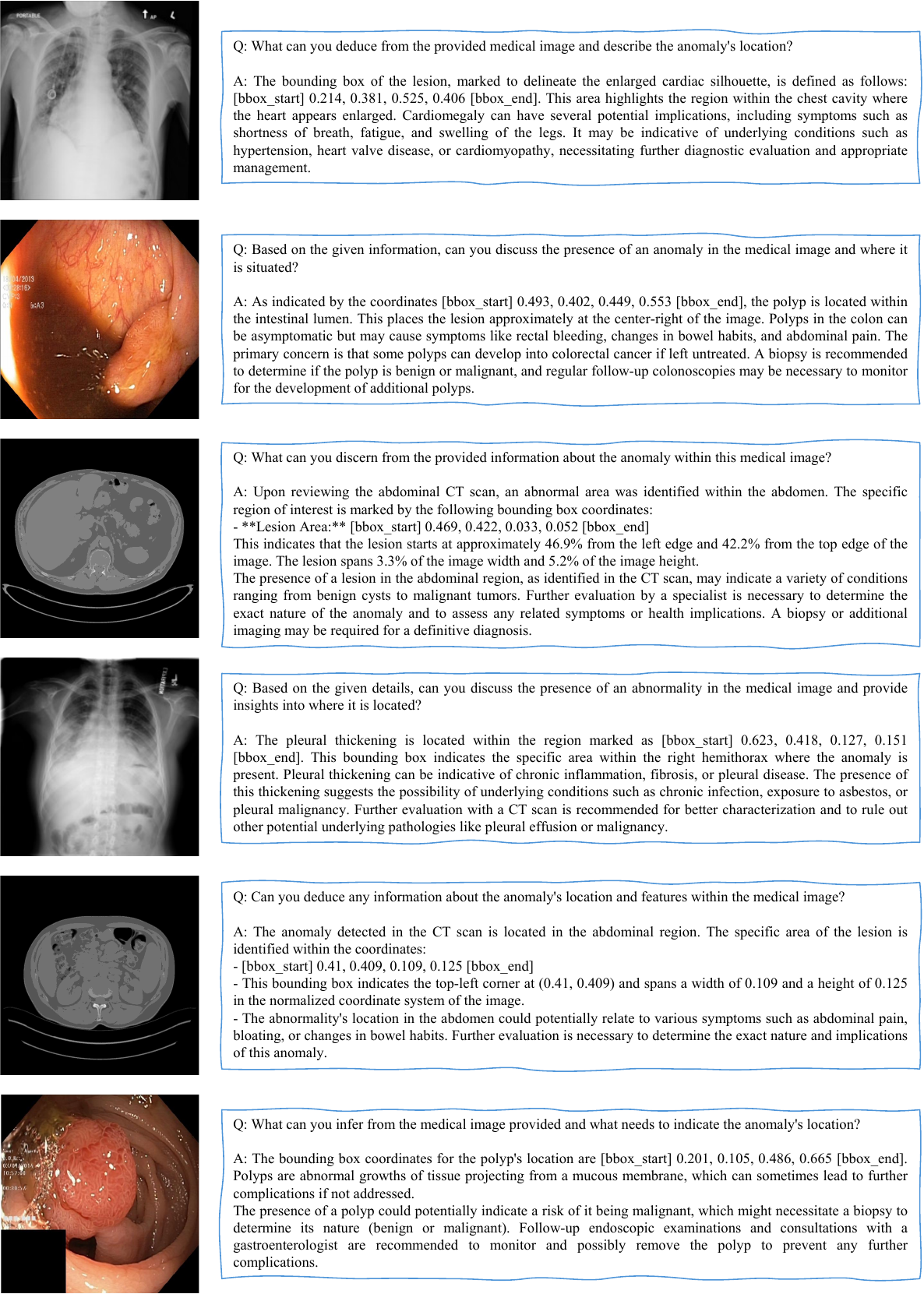}
\caption{\small Examples in MAU datasets.}
\label{fig:examples}
\end{figure*}

\begin{table}[t]\small
\centering
\resizebox{\linewidth}{!}{
\setlength{\tabcolsep}{3pt}
\begin{tabular}{cccrrr}
\toprule
    \bf Dataset & \bf Type & \bf Class & \bf Train & \bf Test& \bf Total\\
\midrule
    DeepLesion & CT & 9 &   1,584 & 396 & 1,980\\
    KidneyStone & CT & 1&   843 & 211 & 1,054\\
    NIH& X-ray & 14 &   787 & 197 & 984\\
    TBX11K & X-ray & 1 &   639 & 160 & 799\\
    KVASIR& Pathology & 1 &   800 & 200 & 1,000 \\
\midrule
    Total& 3 & 26 & 4,653 & 1,164 & 5,817\\
\bottomrule
\end{tabular}}
\caption{\small Specific details of the segments used from datasets.}
\label{tab:data}
\end{table}
The details of dataset are shown in Table~\ref{tab:data}. This dataset encompasses five distinct sub-datasets, namely DeepLesion \cite{DBLP:journals/corr/abs-1710-01766}, KidneyStone \citep{tez_roi_aug_dataset}, NIH \cite{DBLP:conf/cvpr/WangPLLBS17}, TBX11K \cite{DBLP:conf/cvpr/LiuWBWC20}, and KVASIR \cite{DBLP:conf/mmsys/PogorelovRGELJS17}, each originating from diverse sources.
DeepLesion consists of 32,120 axial CT slices featuring eight different types of abnormalities such as lesions in the lungs, abdomen, mediastinum, liver, pelvis, soft tissue, kidneys, and bones.
The KidneyStone dataset contains 1,300 renal CT scans depicting various kidney stones, covering a range of sizes, shapes, and locations within the urinary system.
The NIH dataset includes 112,120 chest X-ray images, covering 14 different pathological categories including, but not limited to, pneumonia, tuberculosis, lung cancer, pulmonary edema, and pulmonary fibrosis.
The TBX11K dataset comprises 11,200 chest X-ray images focused on the classification and detection of tuberculosis (TB), categorizing images into healthy, sick but non-TB, active TB, latent TB, and indeterminate TB.
The KVASIR dataset contains 8,000 endoscopic images of the gastrointestinal (GI) tract, focusing on GI diseases with eight different types of abnormalities, such as polyps, tumors, or other abnormal tissues.
As shown in Figure~\ref{fig:examples}, we present some examples from the MAU dataset.

\section{Implementation Details}\label{app:setting}
In the experiments, we employ MedVInT \cite{zhang2023pmc} to initialize our model.
During the instruction tuning phase, we employed an Adam optimizer \cite{DBLP:journals/corr/KingmaB14} with a learning rate of $1 \times 10^{-5}$, weight decay of 0.01 and a batch size of 128. We train the model on four epochs, utilizing a linear decay strategy for the learning rate.
For the AAR method, we set $\gamma=0.99$, $c_{1}=0.5$, $c_{2}=0.5$, and $c_{3}=0.01$. We used an Adam optimizer with a learning rate of $1 \times 10^{-6}$ and a batch size of 16, training the model for one epoch. To enhance the diversity of sampling responses, we set the number of sampling candidates to eight and decoded the model with a temperature of 0.9 and a probability threshold (p) of 0.9. Our training was conducted on an NVIDIA H800 GPU.

\begin{figure*}[!t]
  \centering
  \includegraphics[width=\linewidth]{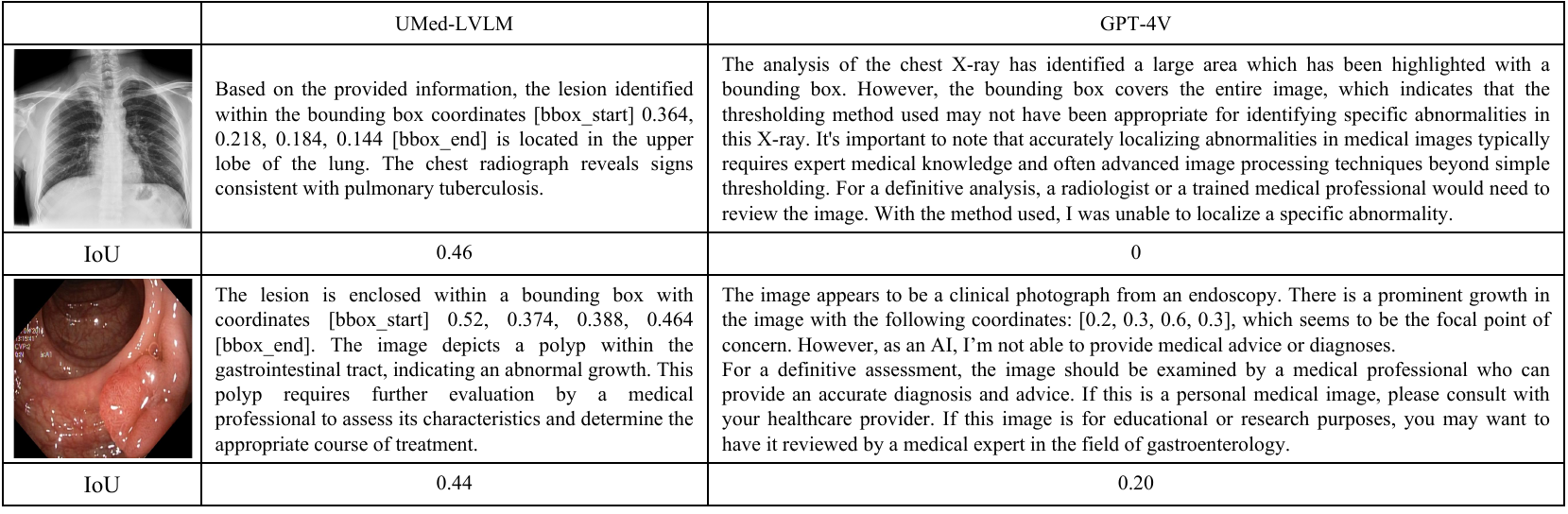}
  \caption{\small The comparison of diagnosis results between UMed-LVLM and GPT-4V on randomly sampled samples.}
  \label{fig:case}
\end{figure*}
\begin{figure*}[!t]
  \centering
  \includegraphics[width=\linewidth]{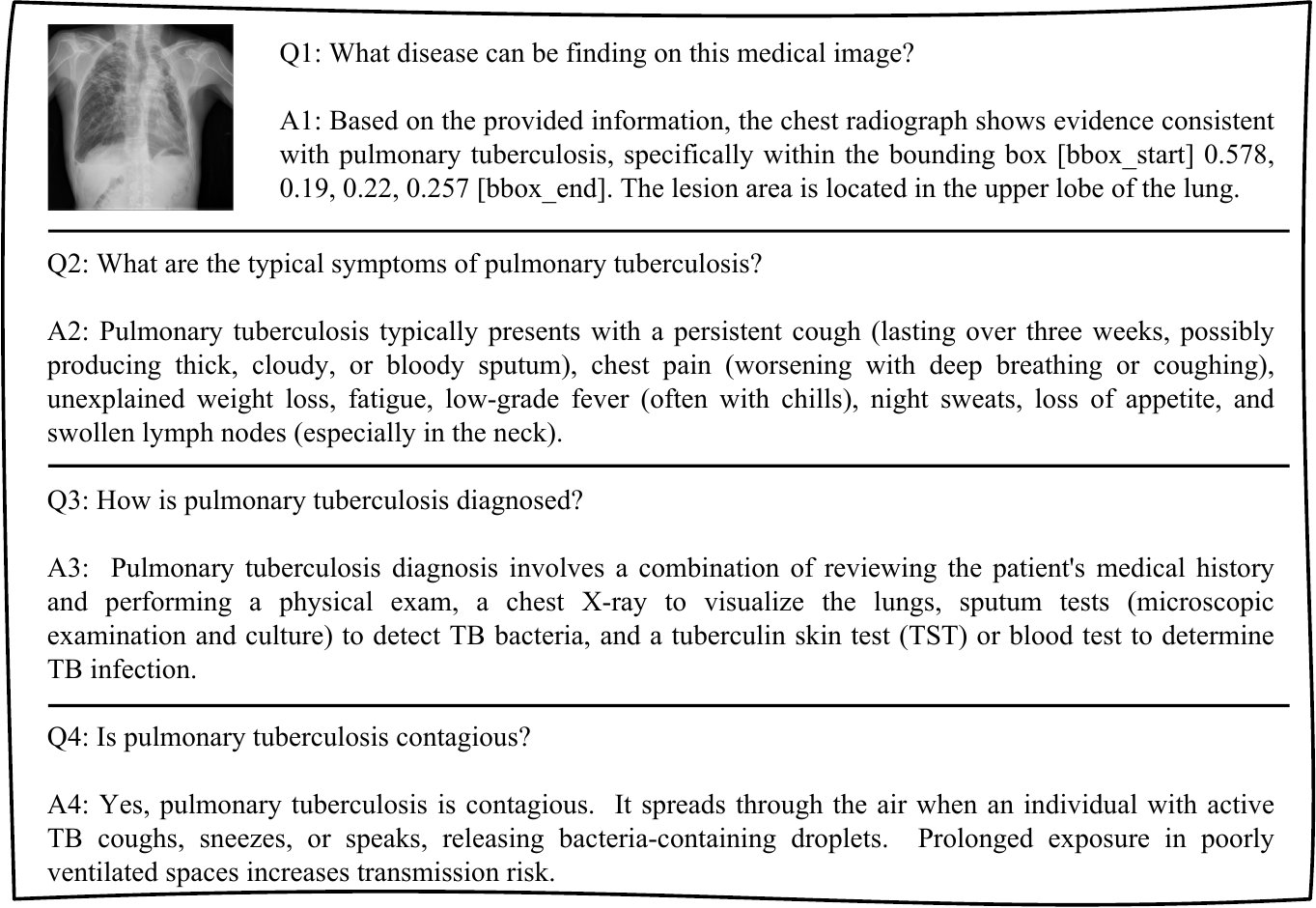}
  \caption{\small Multi-turn conversation from UMed-LVLM.}
  \label{fig:diag}
\end{figure*}

\section{Large Vision-Language Models}\label{app:lvlms}
In this study, we evaluate a variety of large vision-language models (LVLMs) and medical LVLMs (Med-LVLMs), including general-purpose models such as MiniGPT-4~\cite{zhu2023minigpt}, mPLUG-Owl~\cite{ye2023mplug}, LLaVA~\cite{liu2023visual}, Qwen-VL~\cite{bai2023qwen}, and specialized models like Med-LVLMs, including XrayGPT~\cite{thawkar2023xraygpt}, LLaVA-Med~\cite{LLaVA-Med}, Med-Flamingo~\cite{DBLP:conf/ml4h/MoorHWYDLZRR23}, and MedVInt~\cite{DBLP:journals/corr/abs-2305-10415}.
While MiniGPT-4, mPLUG-Owl, LLaVA, and Qwen-VL are designed for general visual tasks, LLaVA-Med, XrayGPT, Med-Flamingo, and MedVInt are specifically optimized for the medical domain. These models generally follow a two-stage training process. In the first stage, they align features using large-scale vision-language datasets, while the second stage fine-tunes the model on instruction-following datasets.
In our experiments, we extended this approach by further training our model on the MAU dataset using Instruction Tuning and ARRL methods, building on the MedVInt framework. For comparison, we also evaluated GPT-4V~\cite{2023GPT4VisionSC}, used in inference mode without additional training due to its closed source.

\begin{table}[ht]\small
\centering
\resizebox{\linewidth}{!}{
\setlength{\tabcolsep}{2pt}
\begin{tabular}{cccccc}
\toprule
(c1, c2) & (0.3, 0.7) & (0.4, 0.6) & (0.5, 0.5) & (0.6, 0.4) & (0.7, 0.3) \\\midrule
MAU (Avg.) & 0.69 & 0.72 & \textbf{0.75} & 0.71 & 0.70 \\
\bottomrule
\end{tabular}}
\caption{Hyperparameter search for Equ.~\ref{equ:reward}.}
\label{tab:Hyperparameter}
\end{table}
\section{Hyperparameter Choices for Equ.~\ref{equ:reward}}
For hyperparameter in Equ.~\ref{equ:reward}, our reference was the source code of PPO\footnote{https://spinningup.openai.com/en/latest/algorithms/ppo.html}.
Furthermore, we conducted hyperparameter search experiments to fine-tune these parameters for our specific environment and task. The results of these experiments, including the specific parameter ranges explored and the final selected values, are detailed in Table~\ref{tab:Hyperparameter}.

\section{Case Study}\label{app:case}
As shown in Figure~\ref{fig:case}, we randomly sample examples to compare our method (i.e., UMed-LVLM) and GPT-4V. For instance, in the first medical image, a chest X-ray, the UMed-LVLM identified a suspicious lesion in the upper lobe of the lung. Despite the IoU score being moderate, the system was able to localize the abnormality, which corresponds to clinical findings consistent with pulmonary tuberculosis. In contrast, GPT-4V failed to localize the abnormality, as it deemed the method it used not suitable for pinpointing specific abnormalities. Similarly, in the gastrointestinal tract image, our UMed-LVLM delineated a polyp with an IoU score of 0.44, suggesting the presence of abnormal growth, whereas GPT-4V with IoU of 0.20. These show our model's proficiency in localizing medical abnormalities. 
We also showcase examples of our UMed-LVLM's capability for multi-turn dialogues in Figure~\ref{fig:diag}.  It demonstrates its potential for interactive medical consultations and diagnostic support.

\end{document}